\documentclass[letterpaper, 10 pt, conference]{ieeeconf}  

\IEEEoverridecommandlockouts                              

\usepackage{graphics} 
\usepackage{epsfig} 
\usepackage{mathptmx} 
\usepackage{times} 
\usepackage{amsmath} 
\usepackage{amssymb}  
\usepackage{booktabs}
\usepackage{caption}
\usepackage{subcaption} 
\usepackage{cite}
\usepackage{rotating}
\usepackage{soul}
\usepackage{wrapfig}

\title{\LARGE \bf
Real-time Estimator of Actuator Control and Health (REACH) on an Eel-Inspired Soft Robot
}

\author{Zhangjingyi Jiang$^{1}$, Myungsun Park$^{2}$, Michael T. Tolley$^{2}$ and Mark Campbell$^{1}$%
\thanks{$^{*}$This work was supported by ONR MURI grant N00014-22-1-2595}
\thanks{$^{1}$Zhangjingyi Jiang, Mark Campbell are with the Department of Mechanical and Aerospace Engineering, Cornell University}
\thanks{$^{2}$Myungsun Park, Michael T. Tolley are with the Department of Mechanical and Aerospace Engineering, University of California San Diego}%
}
\begin{document}
\maketitle
\thispagestyle{empty}
\pagestyle{empty}

\begin{abstract}
An actuator health estimation algorithm for a soft swimming robot that can perform anguilliform swimming is developed. Due to harsh operational environments of underwater robots, and the common degradation of soft robot materials and actuators, accurate estimation of actuator functionality is necessary for robots to perform their missions as well as return to base in the event of actuator degradation and failure. Termed REACH (Real-time Estimator of Actuator Control and Health), the architecture employs a soft robot model, sigma point filter, and a formal statistical hypothesis test to adequately capture the nonlinearities and changes over time. The performance of REACH using three sensor types (GPS, IMU, and Bend Sensor) with one sensor on each actuator is compared, demonstrating that both bend sensor and IMU are adequate choices. Sensor quantity and placement are evaluated for IMU and bend sensor, showing two sensors are sufficient for IMU, whereas three sensors are needed for bend sensor. Three swimming gaits (linear swimming, wide turning, tight turning) are compared, demonstrating that REACH can successfully predict actuator health for all three gaits, with minimal differences in performance. A filter validation method shows the fault estimation algorithm is statistically consistent in finding the correct degradation. The approach is experimentally evaluated using bend sensor data collected from a fish robot, demonstrating that REACH can successfully estimate actuator health with noisy data and variations in manufacturing.
\end{abstract}

\section{Introduction}
Autonomous underwater vehicles (AUVs), as the name suggests, can operate underwater without a human operator \cite{auvDef}. This capability allows AUVs to explore remote areas without requiring to be within a certain range of human operators. AUVs can also be deployed to harsh environments that humans have difficulty reaching \cite{auvUse}. The key to being successful with these missions is the ability to robustly operate over long periods of time without human intervention. This allows missions to avoid complex and high-cost retrieval while enabling longer-range, riskier goals. 

Recent developments in soft robots have led to many new bio-inspired robots that aim to mimic the abilities of their biological counterparts, including water-based robots that operate both on and within water \cite{JacoboRobosoft,waterSoftRobot2}. The flexibility and adaptability of soft robots (continuous deformation, lower drag, lower noise, etc.) make them a good platform for swimming robots \cite{softRobotAd}. A key challenge with soft robots is the compliant materials used are prone to degradation and failure due to changing material properties (Mullins effect) or tears/ruptures \cite{scottH}. Actuator health is difficult to monitor due to the challenges in modeling soft actuators \cite{scottH}. The ability to predict actuator performance allows the robot to adapt accordingly. 

Previous rigid actuator health monitoring methods primarily rely on model-based conditional logic~\cite{PrevWork1,PrevWork3} or machine learning models~\cite{PrevWork21,PrevWork22}, resulting in binary (fail/no-fail) or discrete classification determinations. These are not sufficient for the complexity of soft actuator behavior. 

In this paper, we develop a Real-time Estimator of Actuator Control and Health (REACH), a novel soft robot health monitoring system that can estimate the actuator health of an eel-inspired robot, using an Anguilliform Swimming Soft Robot Simulation Platform (ASSRSimP) \cite{casepaper} as the prediction model. Actuator health is defined as the ratio of actual to desired actuator torque output. An actuator health of zero is full actuator failure, one is full actuator functionality, and above one is over actuation. Actuator health encompasses effects from changes in actuator material (degradation, tears/ruptures), fluid pump function (degradation, failure), and energy supply (power output and stability). We focus on actuators because they are critical to the swimming functionality of soft AUVs, and thus critical to the desired long-term missions. We assume that the soft AUV includes multiple actuators to enable smooth swimming even in the presence of actuator degradation and faults. Our goal is to develop and study (in simulation and experiment) an actuator health estimation approach specific to soft robots that can capture material degradation and faults as quickly and accurately as possible.

The key contributions of this paper include:
\begin{enumerate}
    \item an architecture (sensors, estimator, formal hypothesis test) to estimate the actuator degradation of an eel-inspired soft swimming robot (REACH)
    \item a study of how internal bend sensors compare to inertial sensors for use in REACH
    \item validation of the actuator health estimator using a three-actuator soft robot swimming underwater
\end{enumerate}

\section{Robot Design}
\begin{figure}[hbp!]
  \centering
  \includegraphics[scale=0.25,trim=5cm 2.5cm 10cm 2cm,clip]{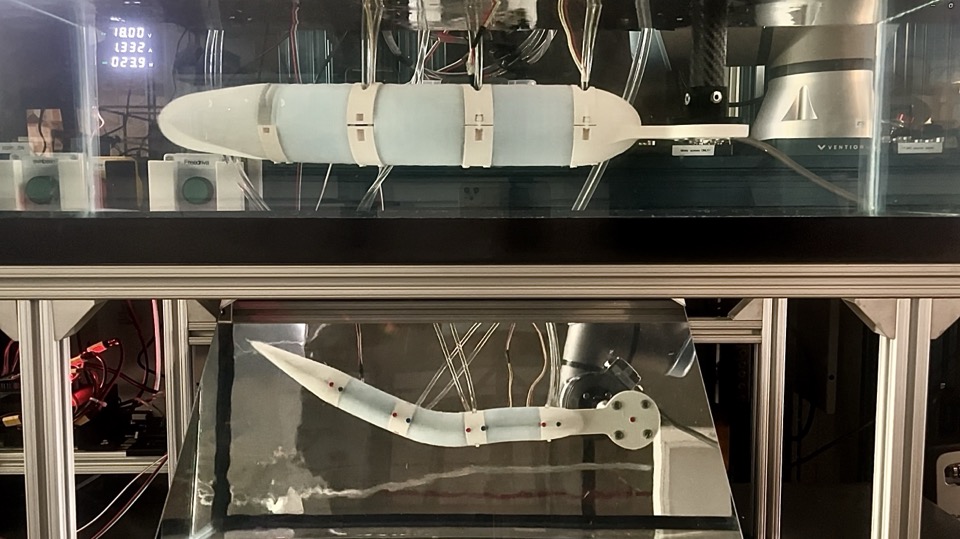}
  \caption{UCSD Robot Fish~\cite{ucsd2025}}
  \label{robotFish}
\end{figure}

We developed a soft robotic fish with anguilliform locomotion~\cite{ucsd2025}. Fig.~\ref{robotFish} shows this robot as it is swimming in a tank. In this section, we summarize key details of the design that are pertinent to the proposed REACH algorithm, namely the materials, sensors, and actuators. 
The robot fish is structured with a rigid head, three self-sensing bending actuators that contain two internal chambers filled with water for each, and a passive tail sequentially connected as a rigid couple. The actuators and the tail were made of Dragon Skin 10 and EcoFlex 30 which are silicone rubbers having 100\% modulus of 186 kPa and 69 kPa, respectively. Each actuator is in the form of an elliptic cylinder with a major axis of 66 mm, a minor axis of 40 mm, and a length of 80 mm. The two internal chambers of each actuator are connected through a hydraulic pump that moves water from one side to the other to create a bending motion by antagonistically expanding/ contracting the volume of the chambers. The bending sensor (Bi-directional flex sensor, Flexpoint Sensor Systems) located on the central axis of the actuator produces a change in the resistance when the actuator bends. This change was measured through a voltage divider circuit. 

\section{Actuator Health Estimation}
\subsection{REACH Algorithm}
The actuator health estimation algorithm (REACH) uses a Sigma Point Kalman Filter (SPKF), due to its ability to handle a wide variety of nonlinearities in estimators, with the Anguilliform Swimming Soft Robot Simulation Platform (ASSRSimP)~\cite{casepaper}, a nonlinear simulation platform for a soft, swimming robot, as its input model.

\subsubsection{ASSRSimP}
Anguilliform Swimming Soft Robot Simulation Platform (ASSRSimP) has two main components: the robot model and the control approach. 

The robot model uses an FEM model of a free-floating elastic beam structure with a nonlinear hydrodynamics model to capture soft robot dynamics, simulating the response of an eel-like fish robot with anguilliform locomotion to internal and external forces. The modularity of the robot model allows for the adjustment of the parameters that define the robot (number of torque inputs/segments, element length, fish shape) as well as key ``soft material'' properties (Young's Modulus, density, shape, etc.) for each element. 

The control approach (Eq.~\ref{cont}) enables swimming, turning, acceleration, and deceleration of the soft swimming robot. 
\begin{equation}
u_{1:n\tau} = [C_A \cdot A_{1:n\tau}] \sin(C_\omega \cdot \omega t + \phi_{1:n\tau}) + C^O_{\tau} \cdot \tau^O_{1:n\tau}
\label{cont}
\end{equation}
where $n_\tau$ is the number of torque inputs, $u_{1:n\tau}$ is the torque input to each actuator, $A_{1:n\tau}$ is the amplitude, $\phi_{1:n\tau}$ is the phase shift, $\tau^O_{1:n\tau}$ is the torque offset for each actuator, $C_A$ is the amplitude coefficient, $C_\omega$ is the frequency coefficient, $C^O_{\tau}$ is the torque offset coefficient, and $\omega$ is the oscillation frequency. 

The amplitude coefficient $C_A$ and the frequency coefficient $C_\omega$ are used to control linear swimming, and the torque offset coefficient $C^O_{\tau}$ is used to control turning. Deceleration (braking) is achieved by reversing the order of the elements in $A_{1:n\tau}$ and $\phi_{1:n\tau}$. Adjustment of actuator health values is achieved by multiplying the actuator torque input $u_i$ of the $i^{th}$ actuator by the health value. The oscillation frequency $\omega$ is consistent across all actuators. $A_{1:n\tau}$ and $\phi_{1:n\tau}$ are preoptimized for each fish design. 

To simulate changes in actuator health, each input torque $u_{i}$ is multiplied by an actuator health $H_i$ as shown:
\begin{equation}
    u_{i}^{\text{REACH}} = H_{i} \cdot u_{i}
    \label{actHealthEq}
\end{equation}
where $i$ is the actuator number, $u_{i}^{\text{REACH}}$ is the torque input for REACH, and $u_{i}$ is the torque input calculated from Eq.~\ref{cont}. $H_{i}$ is the actuator health ranging from zero to one, with zero being non-functional and one being fully operational, 

The actuator health vector will be appended to the state vector as shown in Eq. \ref{updatedState}
\begin{equation}
    X^{\text{REACH}} = \begin{bmatrix}X \\ H\end{bmatrix}
    \label{updatedState}
\end{equation}
where $X$ is the original state vector from ASSRSimP, and $X^{\text{REACH}}$ is the state vector for REACH.

\subsubsection{Sigma Point Kalman Filter (SPKF)}

The SPKF is a nonlinear filtering method that does not require analytical Jacobians -- only the nonlinear dynamics and output equations. As such, a complex, nonlinear hydrodynamic simulator naturally maps well to this filter, given that Jacobians can be challenging to compute. Formally, the SPKF uses weighted samples to calculate means and covariances. 

A set of sigma points (Eq.~\ref{predict1}) is defined at each timestep such that the weighted sample mean and covariance capture the first and second moments of the original distribution. 
\begin{equation}
\begin{split}
S_{k|k} \ & = \text{chol}(P_{k|k}) \\
\chi^0_{k|k} \ & = \hat{x}_{k|k} \\
\chi^{1:2n}_{k|k} & = \hat{x}_{k|k} \cdot 1_n \pm n_\sigma S_{k|k}
\end{split}
\label{predict1}
\end{equation}
where $P$ is the state vector covariance, $S$ is the Cholesky factor of the state vector covariance, $n$ is the number of states, $n_\sigma$ is sigma point scaling factor,$\chi$ is the set of sigma points, and $\hat{x}$ is the predicted state. These points are then passed through the nonlinear dynamics (Eq.~\ref{predict2}) and output functions (Eq.~\ref{update2}). 
\begin{eqnarray}
\chi^{i}_{k+1|k} &=& f(\chi^{i}_{k|k}) \label{predict2} \\
\mathcal{Z}^{i}_{k+1|k} &=& h(\chi^{i}_{k+1|k}) \label{update2}
\end{eqnarray}
where $\mathcal{Z}$ are the sigma point measurements. The resulting points are then used to calculate the weighted sample mean, covariance, and cross-covariance, which are used to calculate the first/second-order moments of the predicted and updated distribution. Thus, only the nonlinear mapping of the dynamics (Eq.~\ref{predict2}) and output (Eq.~\ref{update2})  (and selection of weights) are required for the SPKF. 
\subsection{Filter Validation Method}
The measurement and process noise statistics used in REACH are tuned using a filter validation method to ensure the filter gives statistically significant results. The average normalized innovation squared $\lambda^{KF}_k(N)$ over $N$ time steps is used as a validation test statistic, calculated as shown:
\begin{align}
\hat{z}_{k} &= \sum^{2n}_{i=0} w^i_m \mathcal{Z}^{i}_{k|k-1} \\
v_{k} &= z_{k} - \hat{z}_{k} \\
\chi^2_{Nnz} &= \sum_{k-N}^{k} v_{k}'*inv(S_{k|k})*v_{k} \\
\lambda^{KF}_k(N) &= \frac{\chi^2_{Nnz}}{N} 
\end{align}
where $\hat{z}_{k}$ is the predicted measurement, $w_m$ is the weighting matrix, $v$ is the innovation, $\chi^2_{Nnz}$ is the sum of normalized innovation squared over $N$ timesteps, $S$ is the innovation covariance. 

These values are compared to the two-sided threshold statistic $b_L$ and $b_U$, calculated as:
%
\begin{equation}
b_L = \frac{Inv\{\chi^2_{Nnz}\} (\frac{\alpha}{2})}{N}, \ b_U = \frac{Inv\{\chi^2_{Nnz}\} (1-\frac{\alpha}{2})}{N}
\end{equation}
where $\alpha$ is the false positive rate. 

If $b_L < \lambda^{KF}_k(N) < b_U$, then the filter is consistent. If $\lambda^{KF}_k(N)$ is too high, the filter is biased. If $\lambda^{KF}_k(N)$ is too low, the filter is too conservative. 

Compared to sensor gating, which is simply the rejection of a single outlier measurement, this method averages the normalized innovations squared over a time window of $N$ time steps. Thus, it is used to tune measurement and process noise as well as evaluate filter consistency.

\section{Simulation and Experiment Setup}
\subsection{Simulation}
\subsubsection{Simulation ASSRSimP Model Parameters}
For the use of REACH in simulation, baseline model parameters, as written in Ref. \cite{casepaper}, are used. The simulated fish is a one-meter-long symmetrical elastic robot with a diameter of 0.1 meters. There are five actuators, each 0.2m long, evenly distributed along the fish. The torque inputs are set so the fish robot will swim at full capacity for a short time. Then the actuators degrade/fail at a prescribed time, and the new torque inputs are changed to reflect the new actuator health. Actuator health predictions and variance matrices are generated in real time and analyzed. 

\subsubsection{Measurement Sensors}
The accuracy and timeliness of GPS, IMU, and bend sensor measurements are compared as inputs for actuator health estimation. The covariance of GPS (2D position as latitudinal and longitudinal) measurements is selected to be 1m$^2$ and 1m$^2$ respectively~\cite{gpsNoise}. The covariance of IMU (acceleration, angular velocity) is 0.0061m$^2$/s$^4$ and 0.002rad$^2$/s$^2$ \cite{imuNoise}. The covariance of bend sensor angle measurements is  0.001rad$^2$ \cite{bendSensorNoise}. 
\subsection{Experiment}
\subsubsection{Data Collection}
\begin{figure}[hbp!]
  \centering
  \includegraphics[scale=0.3,trim=3cm 2cm 2cm 14cm,clip]{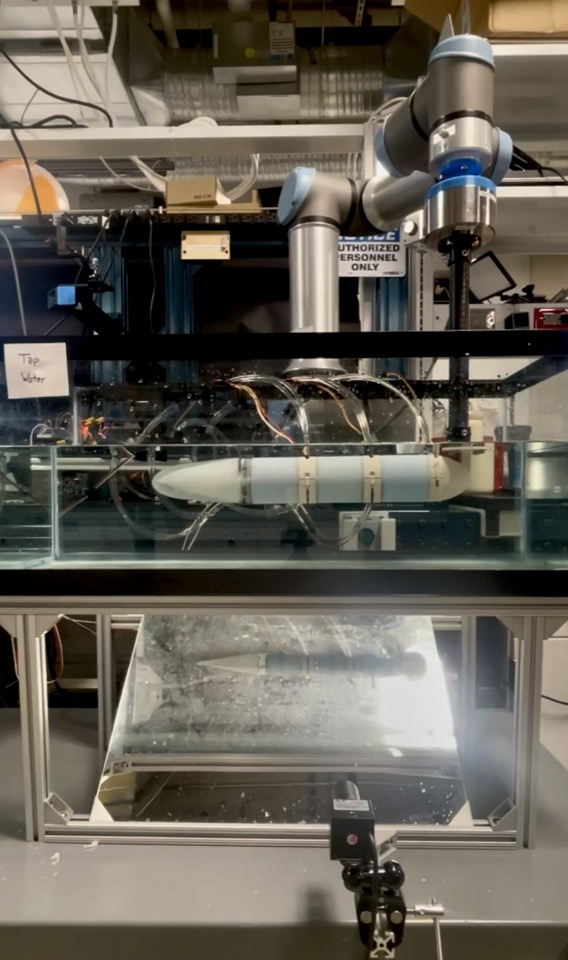}
  \caption{Robot Fish Data Collection Setup}
  \label{robotFishSetup}
\end{figure}
The robot fish was attached to external pumps with flexible tubing. Bend sensors in each actuator were attached to an external microcontroller with electric wires. A camera pointed to a mirror beneath the water tank captured video frames of the eight markers (Fig.~\ref{robotFishSetup}).
In the experiments, body undulation of the robotic fish was generated by giving a square-shaped oscillating flow input to each actuator. Flow rates for full-capacity swimming were set as 575, 345, and 115 ml/min for each actuator (1:3 respectively) to maximize thrust. The data collection setup is shown in Fig.~\ref{robotFishSetup}. Eight markers were attached to the fish as seen in Fig.~\ref{robotFish} (red and black dots on the head, tail, and actuator couplings in the lower image). Video frames showing the fish in a bottom-up view and the bend sensor output were collected at 12 Hz. 
\subsubsection{Data Processing}
\begin{figure}[hbp!]
  \centering
  \includegraphics[scale=0.3,trim=0cm 0cm 0cm 0cm,clip]{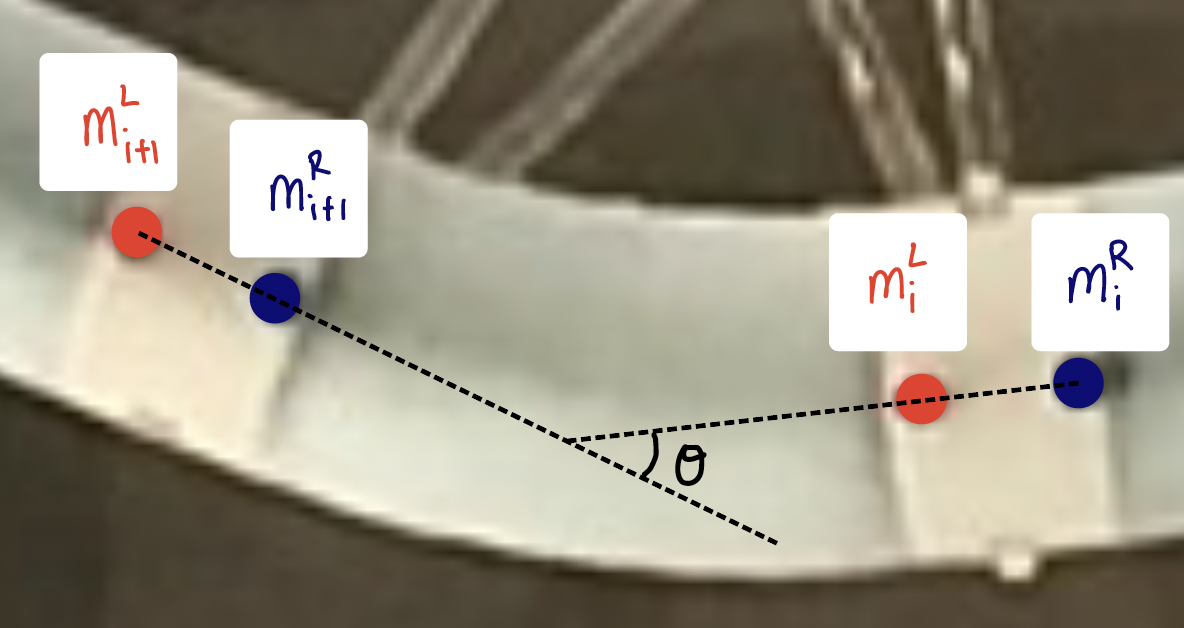}
  \caption{Bend Angle Calculation}
  \label{bendAngleCalculation}
\end{figure}

The four marker positions are extracted from three bend sensor calibration videos using video processing software (Kinovea). Actuator bend angles $\theta$ are calculated for each video frame as shown in Fig.~\ref{bendAngleCalculation} and Eqs.~\ref{bendAngEq1}--\ref{bendAngEq2}. 
\begin{align}
v_i &= m_i^L - m_i^R, \ v_{i+1} = m_{i+1}^L - m_{i+1}^R \label{bendAngEq1} \\
\theta_i &= \mathsf{atan2}(\| v_i \times v_{i+1} \|, v_i \cdot v_{i+1}) \label{bendAngEq2}
\end{align}

where $m_{i/i+1}^{L/R}$ are the four marker positions on either side of the actuator, and $\theta_i$ is the bend angle of the actuator.
A quadratic function (Eq.~\ref{thetaeq}) is used to represent the mapping from the calculated bend angles to the bend angle outputs and is calculated using the {\textsf{fit()}} function in {\textsf{MATLAB}}. This equation is used to convert the bend sensor output to the bend angle for all trials.
\begin{equation} 
\label{thetaeq}
\theta_i = C_1 \cdot x^2_i + C_2 \cdot x_i + C_3
\end{equation}
where $C_{1:3}$ are the coefficients generated from the {\textsf{fit()}} function, $\theta_i$ is the bend angle, and $x_i$ is the bend sensor outputs.
\subsubsection{Experimental ASSRSimP Model Parameters}
The simulation model is adjusted to fit the fish robot to predict actuator health in physical systems. For use on the UCSD robot fish, the simulation model is modified to have seven segments, including three actuators, one head, one tail, and two connections. Each segment has four elements, and the lengths are set to match the robot fish (Table \ref{fishParam}). The Young's Modulus of the head and connection segments is modeled as plastic (Table \ref{fishParam}). The Young's Modulus of actuator segments and tail is modeled as rubber, with exact values adjusted for each actuator's material properties (Section \ref{expValSec}). There are no torque inputs to non-actuator segments.
\begin{table}[htbp]
    \centering
    \caption{Experimental Fish Parameters}
    \label{fishParam}
    \begin{tabular}{@{}cc@{}}
        \toprule
        Variables & Values \\
        \midrule
        Actuator Length $L_\text{actuator} (\text{m})$  & $0.092$ \\
        Head Length $L_\text{head} (\text{m})$  & $0.04$ \\
        Tail Length $L_\text{tail} (\text{m})$ & $0.15$ \\
        Connection Length $L_\text{connection} (\text{m})$ & $0.02$ \\
        Young's Modulus of Non-actuators $E_\text{nonAct} (\text{N/m}^2)$  & $2*10^{9}$ \\
        Young's Modulus of Actuators $E_\text{actuator} (\text{N/m}^2)$  & $10^6$ \\
        \bottomrule
    \end{tabular}
\end{table}

\section{Results and Discussion - Simulation}
In this section, the performance of three sensor types (GPS, IMU, and bend sensor), as well as three swimming motions (linear swimming, tight turning, wide turning), to predict actuator health using REACH are compared using simulation. The consistency of REACH is also tested. A full swimming experiment with the physical robot was conducted, and bend sensor data was collected and used as measurement data for REACH to estimate actuator health values and understand the limitations of the theory. 
\subsection{Performance Metrics}
A successful prediction of actuator health is characterized by both accuracy and timeliness. Accurate predictions allow for more precise decision-making and adjustments. Timely predictions allow for fast response times and greater maneuverability.
In this paper, each simulation was run for ten seconds, with actuator failure occurring after one second. Rise time is used as a measure of timeliness, defined as the time for actuator health prediction to reach within 0.1 of the true actuator health. RMS error is used as a measure of accuracy, defined as the root mean squared (RMS) of the error in actuator health prediction for the last 100 timesteps (one second), calculated using Eq. \ref{rms1}-\ref{rms2}.
\begin{align}
n &= n_k * n_\tau \label{rms1}\\
RMS &= \sqrt{\frac{1}{n}\sum_{k}^{n_k}\sum_{i}^{n_\tau}\left(\hat{H}_i(t_k) - H_i(t_k)\right)^2}
\label{rms2}
\end{align}
where $n$ is the number of measurements, $n_k$ is the number of timesteps, $n_\tau$ is the number of actuators, $\hat{H}_i$ and $H_i$ are the estimated and true health of actuator $i$, and RMS is the root mean squared error.
\subsection{Comparison of Sensor Measurements}
Many sensors can be used to track actuator health. Comparing their effectiveness in simulation is a cost and time-saving method to selecting the sensor type that best meets design requirements. This paper compares the results of the actuator health estimation algorithm using GPS (position), IMU (acceleration, orientation), and bend sensors in all actuators. 

\subsubsection{Filter Validation}
\begin{figure}[hbp!]
  \centering
  \includegraphics[scale=0.35,trim=1cm 0cm 1cm 1cm,clip]{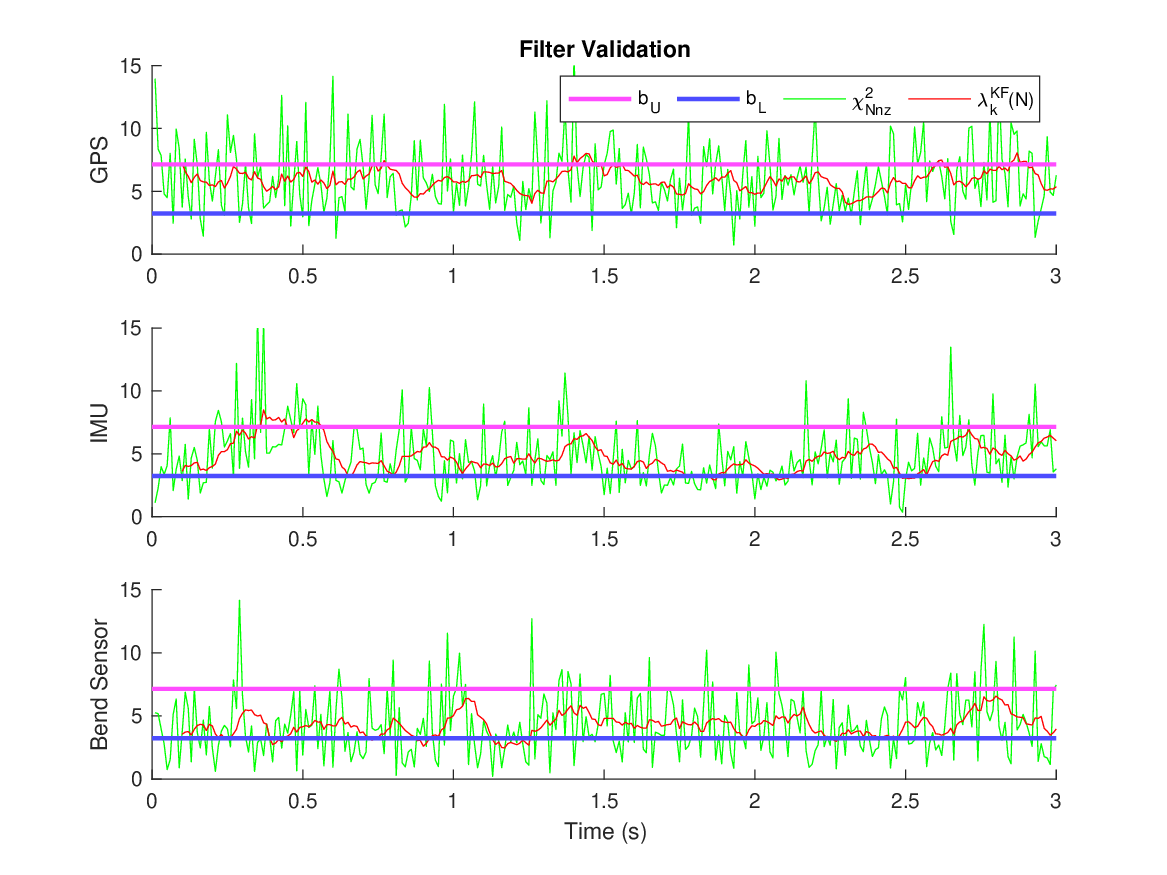}
  \caption{Filter Validation of Actuator Health Estimation Algorithms on Simulation Data}
  \label{filterVal}
\end{figure}

Filter validation is used to tune the process and sensor noise as well as validate the consistency for estimators using GPS, IMU, and bend sensor using a false positive rate $\alpha$ value of $5\%$. Fig. \ref{filterVal} shows the filter validation results after the process and sensor noise tuning. The actuator health estimation algorithm for each estimator is shown to be consistent from $\lambda^{KF}_k(N)$ (red line) falling within $b_L$ and $b_U$ bounds around $95\%$ of the time.
\subsubsection{Actuator Health Estimation - Actuator Failure}
\label{sensActSec}
\begin{figure}[hbp!]
\centering
    \includegraphics[width=0.95\linewidth,trim=0cm 0cm 0cm 1cm,clip]{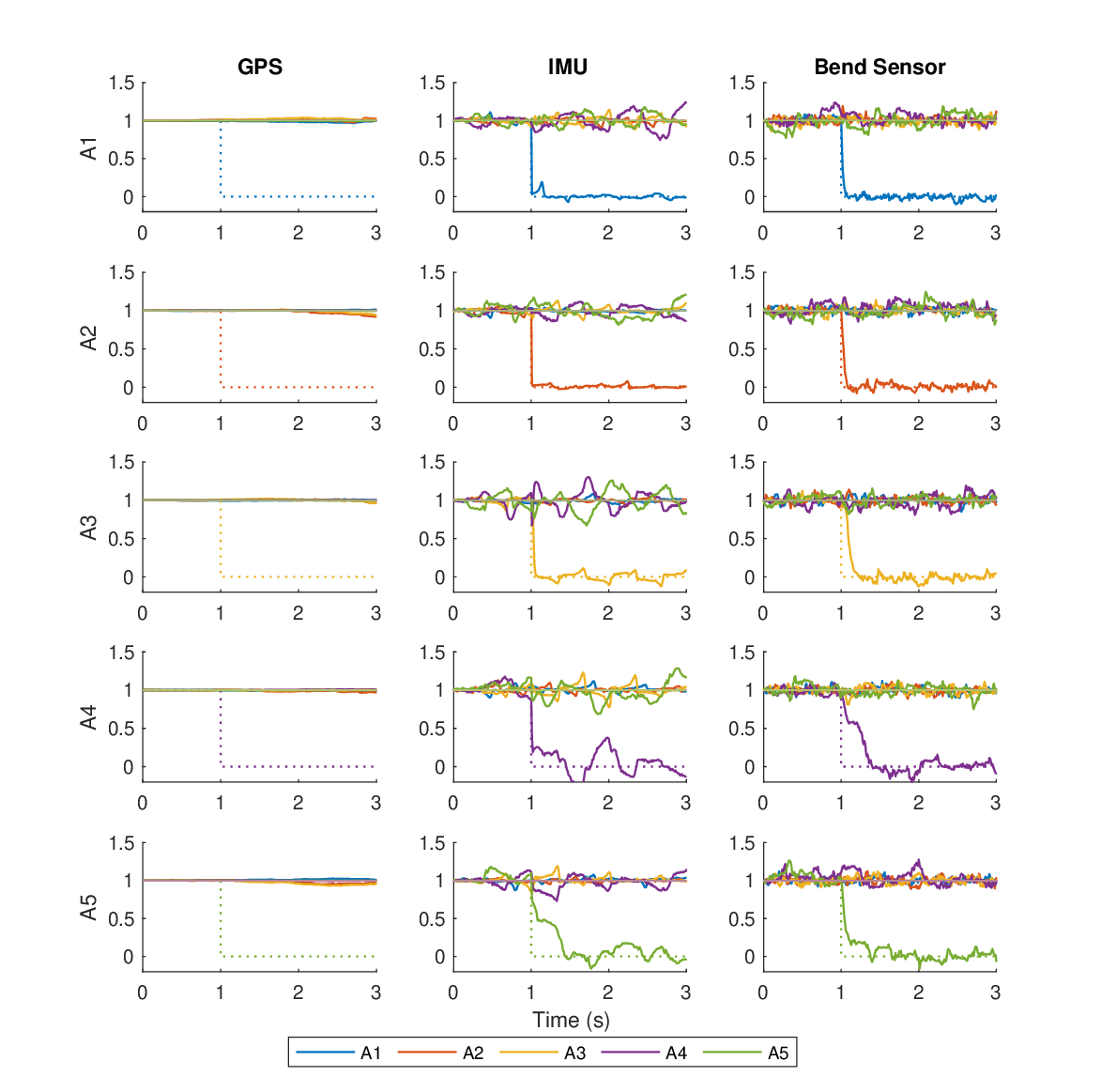}
    \caption{Actuator health estimation using GPS, IMU, and bend sensor of five actuators failures.}
    \label{sensorMsmtComp}
\end{figure}

\begin{figure}[hbp!]
    \centering
    \includegraphics[scale=0.35,trim=0cm 1cm 0cm 0cm,clip]{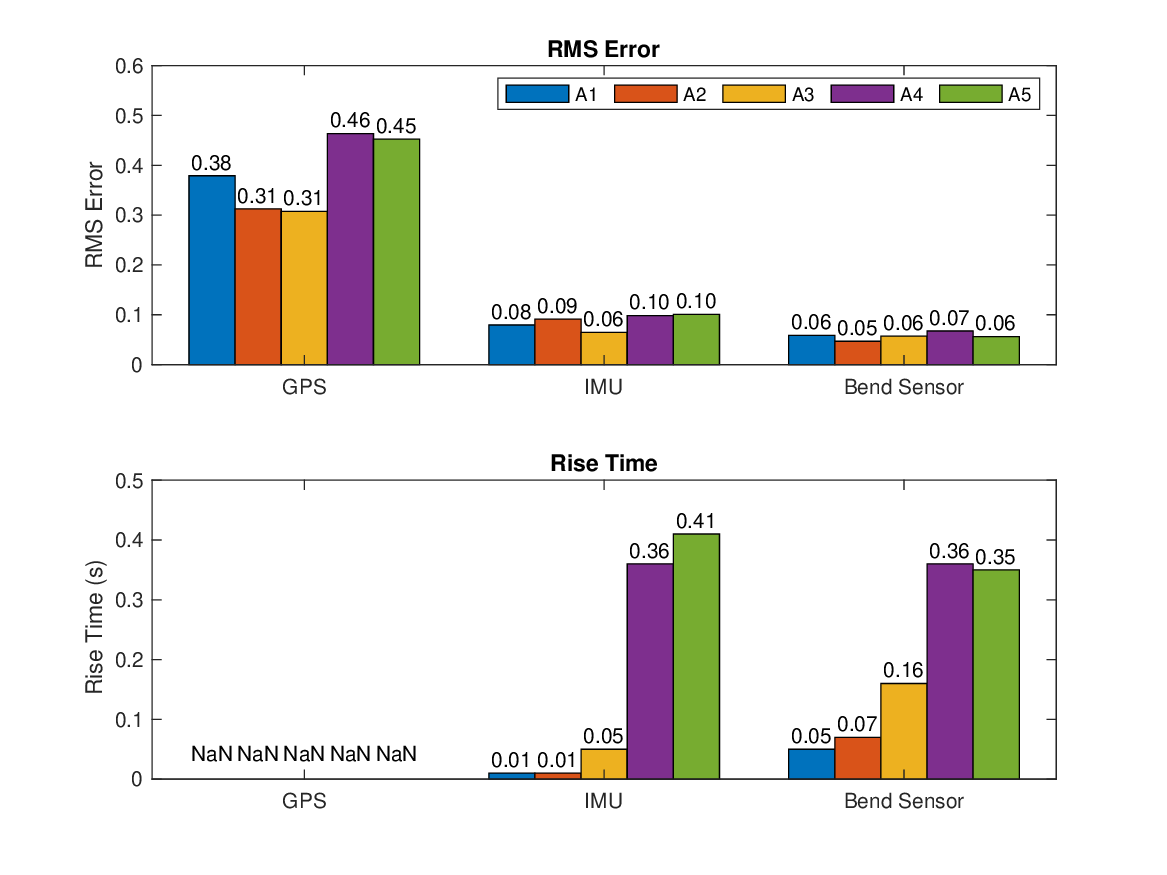}
    \caption{RMS error and rise time of GPS, IMU, and bend sensors in estimating actuator health for each actuator failure case.}
    \label{sensC}
\end{figure}

GPS, IMU, and bend sensor measurements are used to estimate actuator health for each actuator failure case. Fig.~\ref{sensorMsmtComp} shows the actuator health estimations for zero to three seconds, and Fig.~\ref{sensC} shows the RMS error and rise time for each estimation, with A1-A5 representing the failure of actuators 1 through 5 respectively. 
As seen in Fig.~\ref{sensorMsmtComp}-\ref{sensC}, both IMU and bend sensor can detect all actuator failures within three seconds, whereas GPS is unable to detect any failure. Estimated actuator health values for GPS can converge toward the true health after 30 seconds in some simulations but diverge for others. The bend sensor has a lower RMS error for most actuator failure cases and a lower rise time for actuator five failure compared to the IMU. However, for actuator failures in A1 through A4, the bend sensor has a higher rise time than the IMU. Therefore, both IMU and bend sensor are suitable choices for actuator health estimation.
\subsubsection{Actuator Health Estimation - Sensor Placement}
\begin{figure}[hbp!]
    \includegraphics[width=0.99\linewidth,trim=0cm 0cm 0cm 0cm,clip]{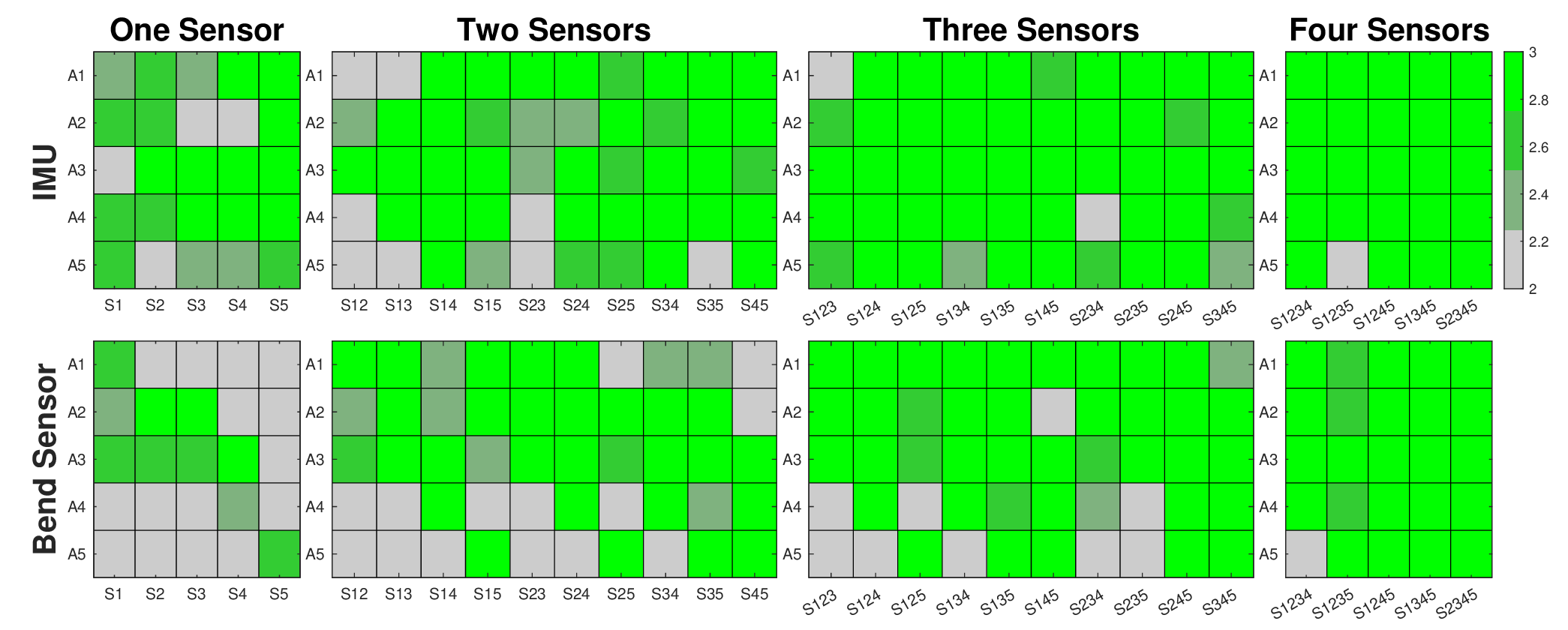}
    \caption{Actuator health estimation using IMU, bend sensor, for one sensor, two sensor, three sensor, and four sensor placement cases.}
    \label{sf1}
\end{figure}

Four sensor quantities with five to ten placement variations each are used to estimate actuator one through five failures respectively. Fig \ref{sf1} shows the estimation performance of each sensor quantity and placement, with the top row for IMU, and the bottom row for bend sensor. Each column group represents one sensor, two sensor, three sensor, and four sensor placement combinations respectively. The vertical axis represents A1 through A5 failure, and the horizontal axis represents sensor placement, with S123 as sensors located at A1, A2, and A3. The performance metric for excellent, good, acceptable, and poor is rise time less than one and RMS error less than 0.15, rise time less than 1.5 and RMS error less than 0.2, rise time less than two and RMS error less than 0.25, and rise time more than two or RMS error more than 0.25 respectively.

As seen in Fig~\ref{sf1}, IMU outperforms bend sensor overall. Bend sensors are more local than IMUs, so they are good at estimating the health of the actuators where they are located. This is evident in all sensor quantity cases, where the bend sensor often failed to estimate actuator health of actuators without sensors. For IMU, two sensors, placed at A1 A4, are sufficient for excellent actuator health estimation. For bend sensor, three sensors, placed at A2 A4 A5, are needed for excellent actuator health estimation.
\subsection{Comparison of Control Algorithms}
Using bend angle as the measurement input, the performance of three control algorithms (Table \ref{ControlSettings1}), in actuator health estimation is compared.
\begin{table}[htbp]
    \centering
    \caption{Control Parameters - Demo 1}
    \label{ControlSettings1}
    \begin{tabular}{@{}cccc@{}}
        \toprule
        Swimming & $C_A$ & $C_\omega$ & $C^O_\tau$ \\
        \midrule
        Linear Swimming & $1$ & $1$ & $0$\\
        Low $C_\tau^O$ Turning$^*$ & $0.8$ & $1$ & $0.3$\\
        High $C_\tau^O$ Turning$^*$ & $0.8$ & $1$ & $0.8$\\
        \bottomrule
        \multicolumn{4}{l}{$^*$Low/High $C_\tau^O$ turn is turning} \\ 
        \multicolumn{4}{l}{with a large/small radius} \\
    \end{tabular}
\end{table}
\begin{figure}[hbp!]
    \centering
    \begin{subfigure}[b]{0.49\linewidth}
        \centering
        \includegraphics[width=0.99\linewidth,trim=1cm 1cm 1cm 1cm,clip]{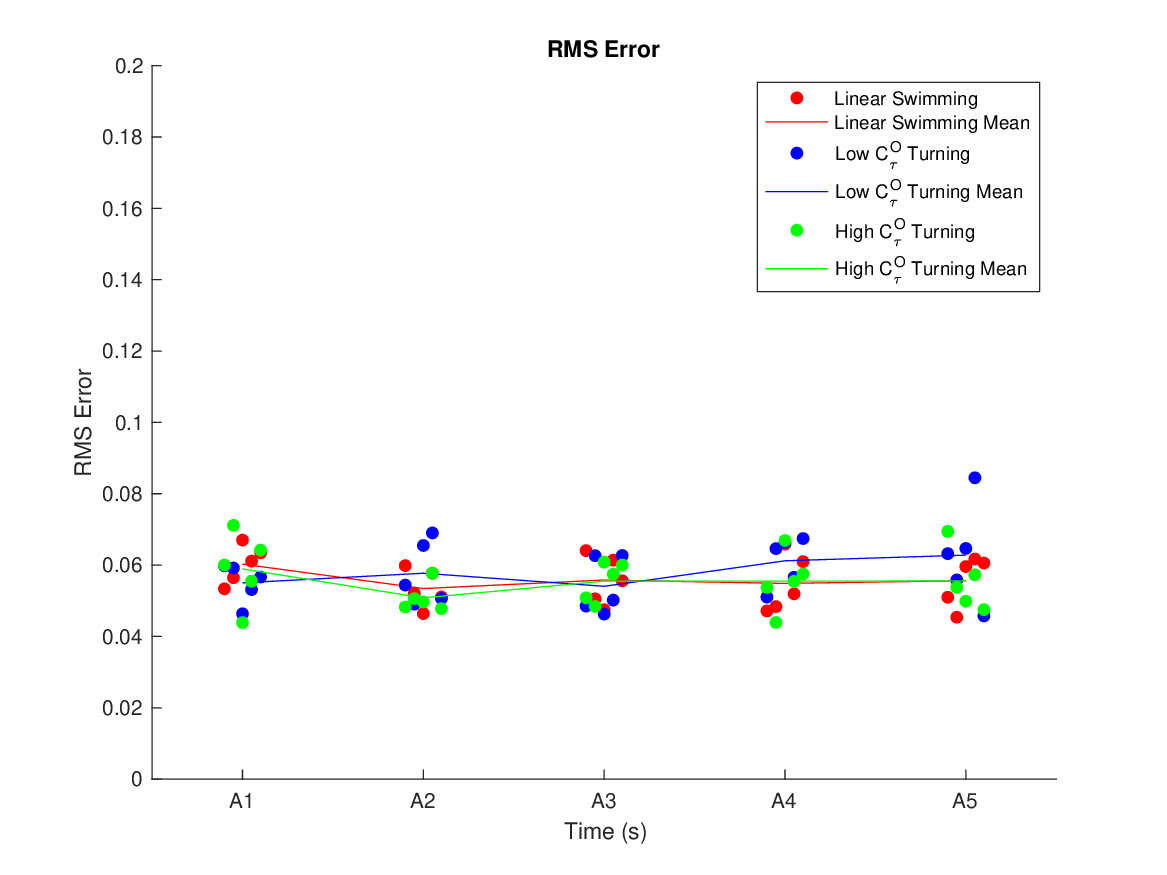}
        \caption{RMS Error}
        \label{RMEerrorCont}
    \end{subfigure}
    \hfill
    \begin{subfigure}[b]{0.49\linewidth}
        \centering
        \includegraphics[width=0.99\linewidth,trim=1cm 1cm 1cm 1cm,clip]{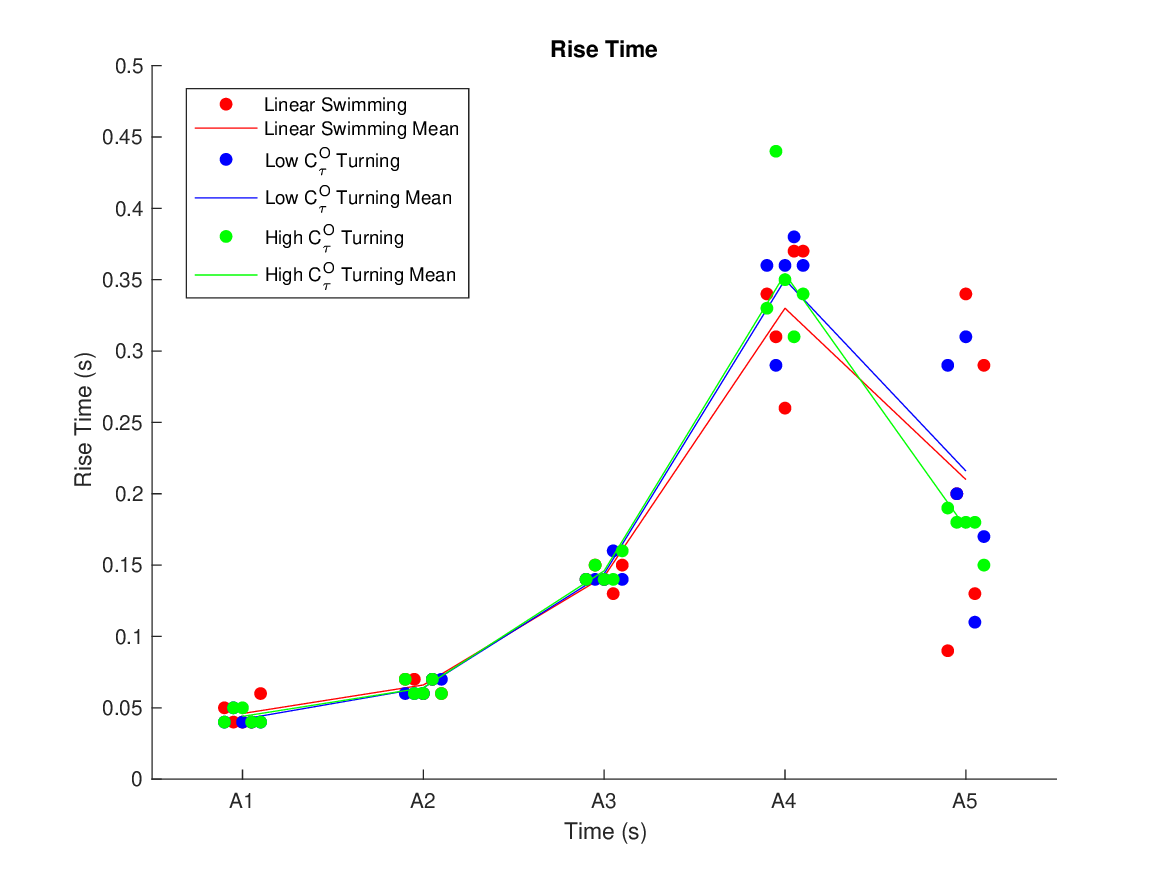}
        \caption{Rise Time (s)}
        \label{risetime}
    \end{subfigure}
    \caption{RMS error and rise time for control algorithms}
    \label{ControlAlg}
\end{figure}
The RMS and rise time of the three control algorithms are simulated five times. Fig.~\ref{ControlAlg} shows no significant difference in the RMS error and rise time between the three control algorithms. Rise time followed the same trend as fig.~\ref{sensC}.

\section{Results and Discussion - Experimental}
\subsection{Experimental Validation}
\label{expValSec}
\begin{figure}[htbp]
    \centering
    \includegraphics[width=0.8\linewidth,trim= 1cm 0cm 0cm 0cm,clip]{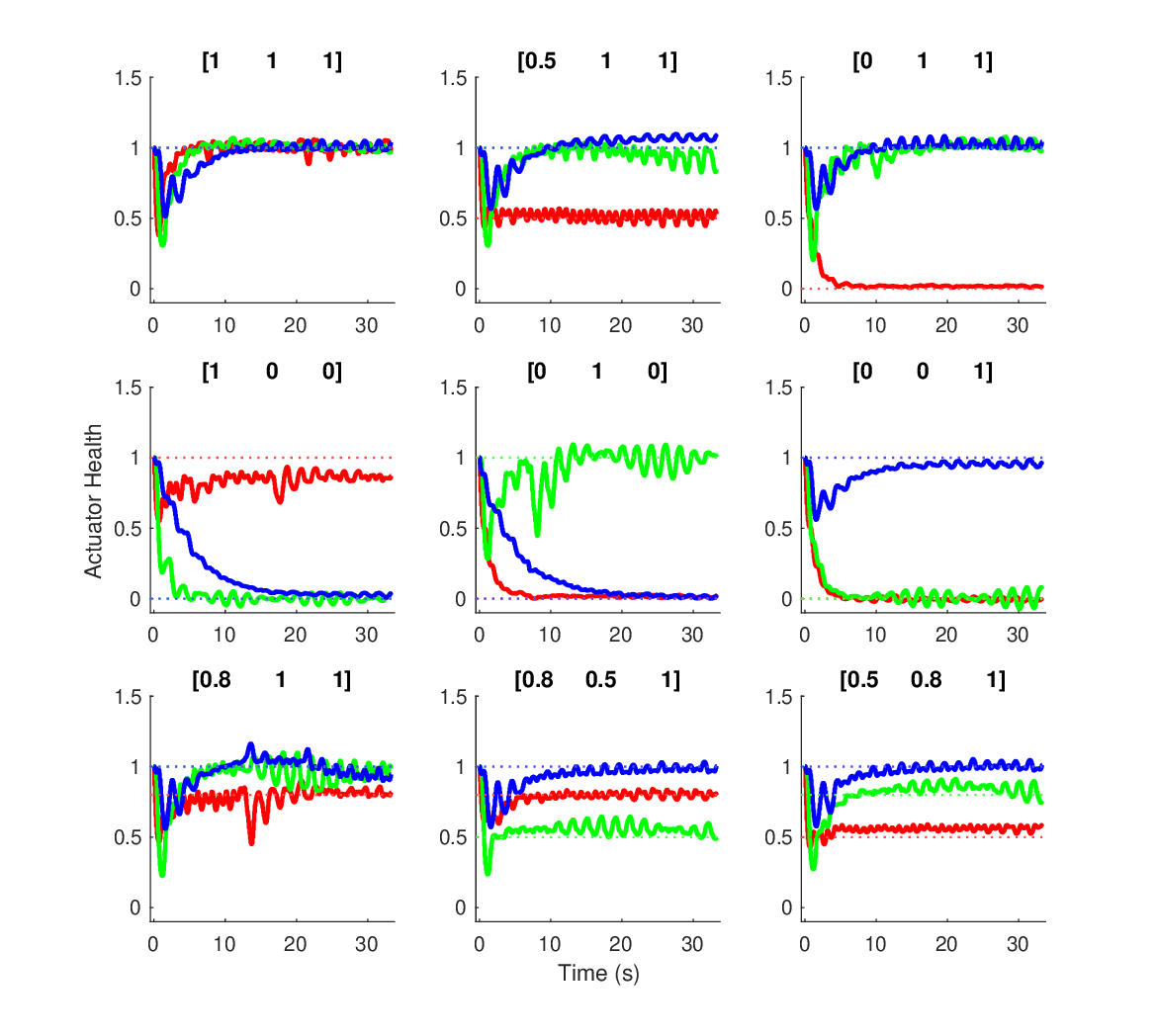}
    \caption{Experimental validation of REACH for nine unique actuator health conditions. A1 Red, A2 Green, A3 Blue}
    \label{exprVal}
\end{figure}

The actuator health estimation algorithm was experimentally validated using data collected from the UCSD robot fish and the adapted simulation model with three actuators. A1 was chosen as the main degradation actuator because it is found to have the most impact on swimming. Due to the difference in stiffness in each actuator, the stiffness of each actuator was scaled to fit the experimental data for full-capacity swimming, with the Young's Modulus for [A1 A2 A3] as [1.14096 0.9121 0.9934]$*10^6 \textrm{N/m}^2$. Those scaled values were then used to estimate actuator health for all other cases.
The pump inputs for each experiment are shown in Table \ref{pumpInput}, where T$_i$ is trial $i$, $[$A1 A2 A3$]$ is the actuator health for actuator $A1$, $A2$, $A3$. Pump inputs at full capacity as 575, 345, and 115 ml/min for each actuator respectively. T1, T2, T3, and T7 show A1 degradation/failure, T4-6 show two actuator failure, and T8-9 show two actuator degradation.
\renewcommand{\arraystretch}{1.5}
\begin{table}[htbp]
    \centering
    \caption{Actuator Health for Each Trial}
    \label{pumpInput}
    \begin{tabular}{@{}c@{ }c@{ }c@{ }c@{ }c@{ }c@{ }c@{ }c@{ }c@{}} \hline 
        T1$^*$ & T2 & T3 & T4 & T5 & T6 & T7 & T8 & T9\\ \hline 
        $[1 \ 1 \ 1]$ & $[.5 \ 1 \ 1]$ & $[0 \ 1 \ 1]$ & $[1 \ 0 \ 0$] & $[0 \ 1 \ 0]$ & $[0 \ 0 \ 1]$ & $[.8 \ 1 \ 1]$ & $[.8 \ .5 \ 1]$ & $[.5 \ .8 \ 1]$\\ \hline 
        \multicolumn{9}{l}{$^*$This trial was used to tune stiffness values}\\
        \hline 
    \end{tabular}
\end{table}
Fig.~\ref{exprVal} shows REACH can accurately estimate actuator health for most cases, with a few having a slight offset. Note in the first few seconds, the estimation error is due to the fluid pump ramping up.
The oscillatory behavior of the actuator health predictions is similar to simulations with a slight time delay in measurements. Therefore, we believe the oscillations are due to a slight time delay in the measurement data. Actuator failure could be predicted well, but exact actuator health values for partially or fully functioning actuators have an offset in some trials. This might be due to the actuator not starting from the neutral position.



\section{Conclusion}
This paper introduces REACH, an actuator health estimation algorithm for an anguilliform swimming soft robot, and simulates it on a one-meter-long five-actuator fish robot. Rise time and RMS error are introduced as performance metrics for accuracy and timeliness. Three sensor types are compared as measurement inputs for REACH. The process and measurement noise for each sensor type are tuned using a filter validation method. The estimation of each actuator failure shows GPS is unsuitable due to long rise time and unreliable behavior. Both IMU and bend sensor has excellent behavior when sensors are used on all five actuators. At least two sensors are needed for excellent prediction using IMU, and three for bend sensor. REACH is demonstrated to be successful for three swimming gaits: linear swimming, wide turning, and tight turning. Swimming experiments were done on a three-actuator soft robot fish with bi-directional bend sensors on each actuator. REACH was experimentally validated using the experimental bend sensor outputs. 

\bibliographystyle{IEEEtran}
\bibliography{Bibliography/Bibliography}

@INPROCEEDINGS{casepaper,
  author={Jiang, Zhangjingyi and Mark Campbell},
  booktitle={2024 IEEE 20th International Conference on Automation Science and Engineering (CASE)}, 
  title={Modeling and Control of an Eel-Inspired Soft Robot for Design Optimization}, 
  year={2024},
}

@INPROCEEDINGS{JacoboRobosoft,
  author={Cervera-Torralba, Jacobo and Kang, Yuxiang and Khan, Eesa M. and Adibnazari, Iman and Tolley, Michael T.},
  booktitle={2024 IEEE 7th International Conference on Soft Robotics (RoboSoft)}, 
  title={Lost-Core Injection Molding of Fluidic Elastomer Actuators for the Fabrication of a Modular Eel-Inspired Soft Robot}, 
  year={2024},
  volume={},
  number={},
  pages={971-976},
  doi={10.1109/RoboSoft60065.2024.10522050}
}

@misc{auvDef,
  author       = {{NOAA Ocean Exploration}},
  title        = {What is an AUV?},
  year         = 2023,
  howpublished = {\url{https://oceanexplorer.noaa.gov/facts/auv.html}},
  note         = {Accessed: 2024-09-15}
}

@article{auvUse,
  title={Bioinspired soft robots for deep-sea exploration},
  author={Li, Guorui and Wong, Tuck-Whye and Shih, Benjamin and Guo, Chunyu and Wang, Luwen and Liu, Jiaqi and Wang, Tao and Liu, Xiaobo and Yan, Jiayao and Wu, Baosheng and others},
  journal={Nature Communications},
  volume={14},
  number={1},
  pages={7097},
  year={2023},
  publisher={Nature Publishing Group UK London}
}

@article{waterSoftRobot2,
  author = {Aubin, C. A. and Choudhury, S. and Jerch, R. and others},
  title = {Electrolytic vascular systems for energy-dense robots},
  journal = {Nature},
  volume = {571},
  pages = {51--57},
  year = {2019},
  doi = {10.1038/s41586-019-1313-1}
}

@INPROCEEDINGS{scottH,
  author={Hamill, Scott and Whitehead, John and Ferenz, Peter and Shepherd, Robert F. and Kress-Gazit, Hadas},
  booktitle={2019 International Conference on Robotics and Automation (ICRA)}, 
  title={Resilient Task Planning and Execution for Reactive Soft Robots}, 
  year={2019},
  volume={},
  number={},
  pages={5148-5154},
  doi={10.1109/ICRA.2019.8794303}
}

@article{gpsNoise,
title = {Unscented kalman filter with process noise covariance estimation for vehicular ins/gps integration system},
journal = {Information Fusion},
volume = {64},
pages = {194-204},
year = {2020},
issn = {1566-2535},
doi = {https://doi.org/10.1016/j.inffus.2020.08.005},
url = {https://www.sciencedirect.com/science/article/pii/S1566253520303286},
author = {Gaoge Hu and Bingbing Gao and Yongmin Zhong and Chengfan Gu},
}

@misc{bendSensorNoise,
    title = {Bend Sensor Technology},
    author = {Flexpoint Sensor Systems},
    year = {n.d.},
    url = {https://flexpoint.com/bend-sensor/},
    note = {Accessed: 2024-10-04}
}

@article{imuNoise,
title = {Static and dynamic validation of inertial measurement units},
journal = {Gait \& Posture},
volume = {57},
pages = {80-84},
year = {2017},
issn = {0966-6362},
doi = {https://doi.org/10.1016/j.gaitpost.2017.05.026},
url = {https://www.sciencedirect.com/science/article/pii/S0966636217302084},
author = {Leah Taylor and Emily Miller and Kenton R. Kaufman}
}

@article{softRobotAd,
  title={Technologies and Applications of Soft Robotics: A Review},
  author={Jyoti Joshi and Avi Raj Singh Manral and Pushpendra Kumar},
  journal={Journal of Graphic Era University},
  year={2023},
  url={https://api.semanticscholar.org/CorpusID:264134013}
}

@Article{PrevWork1,
AUTHOR = {Li, Shiqing and Frey, Michael and Gauterin, Frank},
TITLE = {Model-Based Condition Monitoring of the Sensors and Actuators of an Electric and Automated Vehicle},
JOURNAL = {Sensors},
VOLUME = {23},
YEAR = {2023},
NUMBER = {2},
ARTICLE-NUMBER = {887},
URL = {https://www.mdpi.com/1424-8220/23/2/887},
PubMedID = {36679679},
ISSN = {1424-8220},
DOI = {10.3390/s23020887}
}

@article{PrevWork3,
author = {Gianpietro Di Rito and Francesco Schettini},
title ={Health monitoring of electromechanical flight actuators via position-tracking predictive models},
journal = {Advances in Mechanical Engineering},
volume = {10},
number = {4},
pages = {1687814018768146},
year = {2018},
doi = {10.1177/1687814018768146},
URL = {https://doi.org/10.1177/1687814018768146},
eprint = {https://doi.org/10.1177/1687814018768146}
}

@article{PrevWork21,
  author    = {T. Kovacs and A. Ko},
  title     = {Monitoring Pneumatic Actuators’ Behavior Using Real-World Data Set},
  journal   = {SN Computer Science},
  volume    = {1},
  pages     = {196},
  year      = {2020},
  doi       = {10.1007/s42979-020-00202-2}
}

@inproceedings{PrevWork22,
author = {Ruiz-Cárcel, Cristobal and Starr, Andrew},
year = {2015},
month = {06},
pages = {},
title = {Development of a Novel Condition Monitoring Tool for Linear Actuators}
}

@inproceedings{ucsd2025,
  author    = {M. Park and J. Cervera-Torralba and I. Adibnazari and M. T. Tolley},
  title     = {Analysis of kinematics and propulsion of a self-sensing multi-DoF undulating soft robotic fish},
  booktitle = {2025 IEEE International Conference on Robotics and Automation (ICRA)},
  publisher = {IEEE},
  year      = {2025}
}

\end{document}